# AI-guided deep visual proteomics reveals recurrence-associated morpho-molecular signatures within triple-negative breast cancers

Yesung Cho[1+], Ji Hwan Park[1+], Chan Il Kim[1+], Hyewon Kim[5], Honglan Li[1], Yumin Lee[1], Geongyu Lee[1], Sujeong Hong[1], Seong Min Park[2], Yoonyoung Lee[5], Hee Sool Rho[5], Sumin Lee[5], Amos Chungwon Lee[5], Changhwan Lee[1], Hwanyoung Shim[1], Hyunwook Kim[1], Hyeji Shin[1], Sanha Park[1], Jihoon Yu[1], Yoon Hee Shin[1], Sooheon Kim[1], Hyunjin Park[1,6], Seung Min Park[3], Sangwan Kim[3], Yujung Kim[3], Jinwon Jeong[1], Sung-Im Do[4], Eun-Young Kim[7], Dongmyung Shin[1,2*], Ingu Do[4*], Jongbae Park[1,2,3*]

[1] OmixAI Inc., Seoul, Republic of Korea

[2] Oncocross Co. Ltd., Seoul, Republic of Korea

[3] Department of Basic Biomedical Sciences, Kyung Hee University, Seoul, Republic of Korea

[4] Kangbuk Samsung Hospital, Sungkyunkwan University School of Medicine, Seoul, Republic of Korea

[5] Meteor Biotech Co. Ltd., Seoul, Republic of Korea

[6] College of Pharmacy, Ewha Womans University, Seoul, Republic of Korea

[7] Department of Surgery, Kangbuk Samsung Hospital, Sungkyunkwan University School of Medicine, Seoul, Republic of Korea

[+] These authors contributed equally to this study.

*Corresponding author: Dongmyung Shin, Jongbae Park and Ingu Do

## Abstract

Outcome-informed AI can localize prognostically relevant regions in digital pathology, but directly characterizing the molecular basis of these spatial predictions remains challenging. Here, we developed an outcome-informed, AI-guided deep visual proteomics framework that uses recurrence-risk heatmaps to guide physical tissue sampling and mass spectrometry-based spatial proteomics in triple-negative breast cancer (TNBC). In 156 patients, high- and low-risk regions coexisted within the tumor compartment and showed distinct nuclear and architectural features, indicating intratumoral variation beyond broad tissue identity. Guided by these maps, we physically isolated and profiled 46 AI-defined high- and low-risk tumor regions from two recurrent cases. High-risk regions were enriched for mitotic programs, whereas low-risk regions showed immune- and antigen-presentation-associated programs. We used proteins that consistently differed between high- and low-risk regions to construct a 13-protein composite. The composite showed a significant effect to discriminate highly recurrent patients from less recurrent patients in an expanded TNBC cohort, while the corresponding transcript-based score showed a similar trend in METABRIC. Integrating the protein composite with the H&E-derived risk score improved out-of-bag discrimination, increasing the C-index from 0.679 to 0.739 and the 5-year time-dependent AUC from 0.675 to 0.770. These findings establish outcome-informed H&E risk maps as spatial guides for molecular discovery, extending prognostic AI from prediction toward direct biological interrogation of recurrence-associated tissue states.

## Keywords

Triple-negative breast cancer, Recurrence, Digital pathology, Artificial intelligence, Spatial proteomics, Tumor microenvironment

# Introduction

Hematoxylin and eosin (H&E) – stained whole-slide images (WSIs) capture diverse aspects of tumor morphology and the surrounding microenvironment and remain the foundation of routine histopathological assessment[1,2]. Advances in computational pathology have enabled artificial intelligence (AI) models to extract prognostic information directly from these images[3-7]. In addition to generating patient-level predictions, patch-based models, attention maps, and related spatial outputs can localize associated with predicted clinical risk[3-7]. Such spatial predictions provide a view of where prognostically relevant image features occur. However, localizing prognostic regions on H&E images does not establish the molecular characteristics of the highlighted tissue. A region assigned high risk may reflect increased proliferation, altered immune activity, stromal remodeling, or combinations of these processes, and histologically similar regions may differ substantially at the molecular level[8-10]. Thus, an important challenge is to move from identifying prognostic image regions to experimentally defining the biology that distinguishes them and further implicates spatial heterogeneity in therapeutic development.

To address this gap, prior studies have linked histomorphology with molecular measurements through computational inference and spatial molecular profiling. Computational models can infer bulk or spatial gene-expression patterns from histological images[8-9], while spatial transcriptomics and highly multiplexed imaging can directly characterize molecular variation within tissues[11-14]. At the protein level, highly multiplexed imaging preserves spatial context and enables the characterization of outcome-associated cellular niches in cancer. The QUICHE study, for instance, applied highly multiplexed spatial protein imaging to identify recurrence-associated cellular niches in TNBC[15]. While this modality enables spatially resolved protein characterization, its molecular coverage is determined by the selected marker panel. In contrast, mass spectrometry (MS)-based proteomics enables broad, antibody-independent measurement of thousands of proteins[11-13,16]. In conventional bulk analyses, however, MS-based proteomics averages signals across whole specimens, obscuring their spatial origin[11-14,16]. Combining spatially targeted tissue isolation with MS-based proteomics offers a way to overcome this trade-off, retaining histological context while enabling deep molecular characterization of selected tissue regions.

Deep visual proteomics has demonstrated the value of coupling image analysis, physical tissue isolation, and ultrasensitive MS to connect visually defined cellular phenotypes with their proteomic states[17]. In such approaches, however, sampling regions are generally determined by observable cell identity, morphology, marker expression, or predefined phenotypic classes. Prognostic AI can also be used to define sampling regions according to their association with patient outcome. This is relevant in heterogeneous tumors, where biologically distinct risk states may coexist within the same histological compartment and be obscured by tissue-level averaging[18-19]. An outcome-guided spatial sampling strategy could therefore link prognostic morphology to its underlying molecular features and identify recurrence-associated morpho-molecular signatures, which are not apparent from either image interpretation or bulk molecular measurements alone.

Here, we applied this strategy to triple-negative breast cancer (TNBC) by developing an outcome-informed, AI-guided deep visual proteomics framework. We generated patch-level recurrence-risk maps from H&E images and used matched bulk proteomic profiling to characterize the biological programs associated with the corresponding patient-level risk scores. We then investigated how regional risk was distributed across histological tissue compartments and whether contrasting high- and low-risk states could coexist within tumor tissue itself. The resulting risk maps were subsequently used to select high- and low-risk tumor regions for physical isolation and MS-based spatial proteomic profiling. This design enabled direct comparison of the molecular programs associated with contrasting AI-defined morphologies within the same broad histological compartment, rather than relying on comparisons between predefined tissue classes.

Finally, we investigated whether the molecular features identified in the spatially profiled regions retained prognostic relevance at the patient level. Concordant differences between high- and low-risk tumor regions were used to construct a 13-protein composite, which was evaluated in a larger patient cohort, while the corresponding transcript-based score was examined in the independent METABRIC cohort[20,21]. We further evaluated whether the spatially derived molecular score provided prognostic information complementary to the H&E-derived risk score. Compared with the H&E-derived risk score alone, integrating the spatially derived protein composite increased the out-of-bag C-index from 0.679 to 0.739 and the 5-year time-dependent AUC from 0.675 to 0.770.

Taken together, these findings establish an outcome-informed framework in which H&E-derived risk maps are used not only to localize prognostic image patterns but also to guide molecular interrogation of the corresponding tissue regions. By directing the physical isolation and deep visual proteomic profiling of high- and low-risk regions within the same tumor compartment, this approach links prognostic morphology to spatially localized molecular states. The resulting molecular signals can then be evaluated at the patient level and integrated with H&E-derived risk, providing a route from prognostic prediction to experimentally grounded molecular discovery. Extending this framework to longitudinal specimens and integrating of how spatially heterogeneous tumor states evolve during progression and recurrence.

## Results

### Clinicopathological characteristics of TNBC cohorts

We included 156 patients with TNBC who had available H&E-stained WSIs and recurrence follow-up data. We divided the cohort into a development set of 107 patients and a test set of 49 patients (**Table 1** and **Supplementary Table S1**). Recurrence occurred in 13 patients in the development set (12.1%) and 7 patients in the test set (14.3%), with no significant difference between the sets. Between the two sets, age, time to recurrence, histologic type, histologic grade, lymphovascular invasion, and N stage were comparable

### AI analysis of TNBC H&E WSIs

We developed a patch-based AI framework to generate recurrence-associated risk scores (RRSs) from H&E-stained WSIs of TNBC (**Fig. 1a**). Following patch extraction and tissue-quality filtering, a weakly supervised classifier generated a patch-level RRS for each retained patch. Patch-level RRSs were remapped to their original WSI coordinates to generate spatial recurrence-risk heatmaps.

We examined the distribution of patch-level RRSs by patient-level recurrence status in the test cohort (**Fig. 1b**). Patches from patients with recurrence were more frequently represented in the upper RRS bins and least represented in the 0.0–0.2 bin. We subsequently compared the average fractions of patches assigned to each RRS bin between the recurrent and non-recurrent groups (**Fig. 1c**). Patients without recurrence had a higher mean patch fraction in the lowest-risk bin, whereas those with recurrence had higher mean fractions in the 0.6–0.8 and 0.8–1.0 bins. Individual WSIs also contained mixtures of low- and high-scoring patches, demonstrating substantial within-slide heterogeneity in patch-level RRS values (**Supplementary Fig. 1**).

Based on these findings, we further investigated whether preserving the patch-level RRS distribution and selectively incorporating high-scoring patches improved patient-level risk aggregation. Thus, we compared mean- and histogram-based aggregation using either all retained patches or the highest-scoring patches. Following model selection, the final Hist. + Top-k configuration consisted of the 10 highest-scoring patches, a 20-bin histogram representation, and Lasso regression (**Supplementary Tables S2–S3**). Among the evaluated aggregation strategies, Hist. + Top-k showed the best discrimination in the test cohort, supporting that the aforementioned patch-level RRS signal carries recurrence-associated information at the patient level (**Fig. 1d**). The aggregation also achieved comparable discriminative performance to representative WSI-level weakly supervised pooling methods (ABMIL[22], CLAM[4], and TransMIL[23]) evaluated under the same framework (**Fig. 1e** and **Supplementary Table S4**).

**TNBC bulk proteomics analysis**

To investigate the biological characteristics associated with the patient-level RRSs generated by the H&E-based AI model, we performed bulk proteomic profiling of samples from the test cohort (n = 49; **Fig. 2a** and **Supplementary Table S5**). After quality control and preprocessing, we classified 41 patients with evaluable proteomic data into the high-risk (n = 8; 19.5%), mid-risk (n = 9; 22.0%), or low-risk (n = 24; 58.5%) group according to their AI-derived RRSs. Data-independent acquisition (DIA)-based proteomic analysis[24] identified 8,876 proteins corresponding to 126,005 unique peptides, of which 6,609 were retained for downstream analysis.

We characterized the proteomic state associated with each risk group by comparing each group with the remaining patients by differential protein expression analysis. We identified 118, 44, and 77 preferentially upregulated proteins in the high-, mid-, and low-risk groups, respectively (**Fig. 2b** and **Supplementary Table S6**). We then performed functional enrichment analysis to explore the biological processes associated with each risk group. Distinct functional characteristics were observed among the three risk groups (**Fig. 2c** and **Supplementary Table S7**). DEPs upregulated in the high-risk group were predominantly enriched in cell-cycle and genome-maintenance programs, including cell cycle regulation, DNA integrity checkpoint signaling, mitotic nuclear division, and mitotic sister chromatid segregation. Representative high-up proteins, including CDK2, TRIP13, UHRF1, SMC2, and RCC1, are involved in active cell cycle progression, mitotic chromosome segregation, and genomic integrity maintenance, suggesting that the high-risk group reflects a tumor cell–intrinsic proliferative phenotype. In the mid-risk group, mid-up DEPs were mainly associated with fibroblast proliferation, FGFR signaling regulation, cellular responses to reactive oxygen species, and blood vessel endothelial cell migration. This pattern suggests that the mid-risk group is characterized by a stromal-like remodeling phenotype with angiogenic activity within an oxidative TME, distinct from the other biological states. In contrast, DEPs upregulated in the low-risk group were strongly enriched for immune-related biological processes, including immune system process, MHC protein complex assembly, and lymphocyte activation. Key proteins such as CD3E, LCK, PIK3CD, and HLA-DRB5 are involved in antigen presentation and T cell–mediated immune activation, suggesting that the low-risk group is characterized by an immune-enriched TME with increased lymphocyte activation.

To validate these group-associated molecular phenotypes at the cellular composition level, we performed ESTIMATE[25] deconvolution analysis (**Fig. 2d–2e** and **Supplementary Table S8**). As in the enrichment analysis, the low-risk group showed significantly higher immune scores than the other groups, supporting an immune-enriched proteomic state. The high-risk group tended toward higher tumor purity, consistent with a tumor cell–intrinsic proliferative phenotype. Conversely, stromal scores were not significantly elevated in the mid-risk group, suggesting that the mid-up program reflects localized remodeling and growth factor/angiogenic signaling rather than a broad increase in overall stromal content.

### Morphological characterization of high- and low-risk patches

Given the distinct proteomic programs associated with the AI-derived patient-level risk groups, we examined the relationship between patch-level RRSs and histological tissue compartments. Using a TIGER[26]-based tissue classifier, each retained patch was assigned to one of four compartments: tumor, tumor-associated stroma (TAS), necrosis, or immune. The resulting compartment distributions were comparable between the development and test cohorts (**Supplementary Fig. 2a**). In both cohorts, TAS was the predominant compartment, followed by tumor and necrosis, whereas immune patches were rare and accounted for <1% of all retained patches. Following DINO[27]-based normalization, patch features were projected into a UMAP[28] space to visualize the alignment between tissue compartment identity and the RRS landscape (**Fig. 3a–3b** and **Supplementary Fig. 2b**). The UMAP[28] showed compartment-associated organization, with different tissue types showing distinct spatial tendencies within a heterogeneous RRS landscape. Following these observations, we characterized the tissuecompartment composition within each patch-level RRS bin and compared the overall RRS distributions among compartments (**Fig. 3c–3d**). Despite their low overall abundance, immune patches were predominantly confined to the lower-risk bins and were nearly absent from the two highest-risk bins (0.6–1.0). Consistently, immune patches showed lower overall RRSs than the other tissue compartments, aligning with the immune-enriched phenotype observed in the low-risk proteomic group. TAS remained the predominant compartment across all RRS bins and was most abundant in the intermediate bin (0.4–0.6), whereas tumor patches accounted for larger relative fractions in the highest (0.8–1.0) and lowest (0.0–0.2) bins than in the intermediate bins. (**Fig. 3c**). Comparison of the overall RRS distributions across tissue compartments showed that TAS had the highest median RRS, exceeding that of the tumor compartment, whereas tumor exhibited a broader distribution extending toward both low- and high-risk bins (**Fig. 3d**). This broad distribution indicates substantial within-tumor heterogeneity, with measures of central tendency potentially obscuring distinct low- and high-risk patch subsets within the tumor compartment.

Subsequently, we assessed whether patch-level RRSs followed a unimodal distribution within each tissue compartment using Hartigan's dip test[29]. RRS distributions were non-unimodal in tumor, TAS, and necrosis patches; however, unimodality was not rejected for immune patches (**Fig. 3e** and **Supplementary Fig. 2c–2e**). This distributional heterogeneity was particularly evident in the tumor compartment. Among patches in the top RRS quartile, tumor patches across the tissue compartments had the highest RRSs (**Fig. 3f**). Conversely, tumor patches showed the lowest score range among patches in the bottom quartile (**Fig. 3g**). Thus, while bulk proteomic analysis primarily highlighted a proliferative tumor-associated program in the high-risk group, patch-level image analysis further supported the presence of a low-risk subset within the tumor compartment. Together, these findings demonstrated substantial within-tumor heterogeneity, with tumor patches encompassing both extremes of the RRS spectrum rather than being uniformly associated with high risk.

To investigate the histological features underlying the observed patch-level risk heterogeneity, we selected the 10 highest- and 10 lowest-scoring patches within each tissue compartment and found distinct histological patterns across high- and low-risk patches within each tissue compartment via board certified histopathological review (**Fig. 3h** and **Supplementary Fig. 2f–2h**). Within the TAS compartment, high-risk patches tended to exhibit denser, more sclerotic stroma than low-risk patches (**Supplementary Fig. 2f**). In contrast, several low-risk patches contained little conspicuous stroma and were composed predominantly of tumor cells. High-risk patches in the immune compartment showed a relatively scattered distribution of immune cells, whereas low-risk patches tended to exhibit aggregated immune-cell patterns (**Supplementary Fig. 2g**). Notably, the tumor compartment showed morphological differences, with patches at both extremes of the RRS distribution (**Fig. 3h** and **Supplementary Fig. 2h**). High-risk tumor patches also showed more irregular nuclear contours than low-risk ones. Conversely, low-risk tumor patches contained tumor cells with relatively rounder nuclei, and several representative low-risk patches showed glandular architecture. These nuclear and architectural differences were broadly consistent with the proliferative phenotype identified in the high-risk proteomic group. Collectively, these observations indicate that the AI-derived patch-level RRS captured histological variation beyond tissue compartment identity, revealing morphologically distinct high- and low-risk subsets within the tumor compartment. Subsequently, spatial proteomic profiling of AI-defined high- and low-risk tumor regions was performed to determine the association between these contrasting tumor morphologies and distinct molecular programs.

**Spatial proteomic profiling of high- and low-risk patches**

We performed spatial proteomics analysis focused on AI-defined high- and low-risk tumor patches to explore whether the morphological heterogeneity observed within the tumor compartment reflects spatially localized proteomic differences, (**Fig. 4a**). Two cases were selected among recurrence cases in the test cohort for whom adjacent formalin-fixed paraffin-embedded (FFPE) sections were available: P1, which showed the greatest within-slide variance in patch-level RRSs, and P2, which had the highest overall patient-level RRS (**Fig. 4b**). From the selected case slides, patches were manually selected based on the AI-derived patch RRS and tissue compartments and subsequently categorized as tumor high-risk ($T^{high}$*)* or low-risk ($T^{low}$) patches. Overall, 58 tumor regions were profiled for spatial proteomics analysis (**Fig. 4c, Supplementary Fig 3a–3b**). The selected patches were isolated using a SLACS platform[30], followed by DIA[24]-based MS proteomic analysis. Because each individual patch yielded a limited number of detectable proteins, each spatial proteomics spot was generated by combining two to six neighboring patches and then analyzed as single region, thereby ensuring sufficient proteomic depth (**Supplementary Fig. 3c**). After preprocessing, 46 tumor regions were retained for downstream analyses, including 19 $T^{high}$ (P1 = 6, P2 = 13) and 27 $T^{low}$ (P1 = 12, P2 = 15), with 5,572 quantified proteins overall (**Supplementary Table S9**).

We next compared high- and low-risk tumor regions within each case to determine whether the AI-derived patch-level RRS captured localized molecular heterogeneity. Overall, 602 (340 high-up and 262 low-up) and 113 (60 high-up and 53 low-up) DEPs were identified in P1 and P2, respectively (**Supplementary Table S10**). Only 21 DEPs overlapped between the two patients, of which 19 showed concordant regulation, including 9 commonly high-up and 10 commonly low-up proteins (**Supplementary Fig. 3d**). The limited protein-level overlap indicated substantial interpatient molecular heterogeneity. Although the overlap at the individual protein level was limited, the two patients showed concordant process-level differences between high- and low-risk tumor regions (**Fig. 4d** and **Supplementary Table S11**). High-risk regions in both patients were enriched in mitotic programs, consistent with the proliferative phenotype observed on proteomic analysis. Low-risk regions showed immune- and antigen-presentation–associated programs within tumor-predominant areas, aligning with the immune-enriched phenotype of the low-risk bulk group. In P1, high-risk regions also showed extracellular matrix organization, collagen metabolic processes, and fibroblast migration, indicating a patient-specific stromal and extracellular matrix-remodeling component.

To distill these spatial proteomic findings into a focused set of candidates suitable for broader evaluation, we examined the overlapping DEPs with concordant directions of regulation between the two patients. The concordant high-risk proteins included CDK12, EPPK1, KRT34, MYH11, NFATC2IP, P4HA1, POU2F1, RPS15, and YAP1, whereas the concordant low-risk proteins included ARHGDIB, CEP41, GET3, HELZ2, LAP3, LCP1, PSME2, SERPINB9, TYMP, and WAS (**Fig. 4e**). Although overlap at the individual protein level was limited, these concordantly regulated proteins represent a shared molecular contrast between AI-defined high- and low-risk tumor regions. Given the limited scalability of spatial proteomic profiling for large-cohort applications, we subsequently translated these spatial candidates into a bulk tissue–level readout by quantifying the corresponding peptide precursors using DIA-MS[24], enabling integration with the H&E-based AI RRS for patient-level risk assessment.

## Validation using spatial proteomics-derived markers and H&E AI RRS

We evaluated whether spatial tumor signatures provided recurrence information beyond that captured by H&E-based risk prediction in an expanded cohort of 96 patients with TNBC, comprising the original test cohort (n = 49) and 47 additional independent patients with matched FFPE H&E whole-slide images and corresponding FFPE tissue specimens. For each patient, a patient-level image-based RRS was generated using the H&E-based AI model, while the matched bulk FFPE tissue was subjected to DIA-MS[26] analysis (**Fig. 5a**). Using targeted post-acquisition analysis of the DIA-MS[26] data, we quantified 33 precursors representing 13 spatial tumor signatures that were consistently detected in both the spatial and bulk test cohort datasets (**Supplementary Fig. S4a** and **Supplementary Table S12**).

We next investigated the prognostic value of tumor-derived signatures as a standalone molecular score. We derived compartment-informed composite scores: a tumor composite comprising 13 proteins (EPPK1, MYH11, P4HA1, RPS15, and YAP1 upregulated in high-risk regions; ARHGDIB, GET3, LAP3, LCP1, PSME2, SERPINB9, TYMP, and WAS downregulated). In the expanded cohort, higher composite scores tended to be associated with poorer recurrence-free survival (**Fig. 5b**). We then evaluated the same 13-marker framework using transcriptomic data from METABRIC[20,21], one of the largest independent TNBC cohorts with recurrence follow-up (**Fig. 5c**). The composite stratified patients into groups with distinct recurrence-free survival, with higher scores associated with poorer outcomes. Most individual markers retained the expected direction of association, supporting the reproducibility of the spatially derived tumor signature in an external TNBC population (**Fig. 5d**). Collectively, these findings indicate that the spatially derived 13-marker signature can stratify patients with TNBC by recurrence outcome.

Finally, we assessed whether the tumor composite provided prognostic information beyond the H&E-derived RRS in the expanded cohort out-of-bag bootstrap evaluation (OOB). As a standalone predictor, the tumor composite showed modest recurrence discrimination (OOB AUC, 0.683; OOB C-index, 0.643), comparable to the H&E RRS (OOB AUC, 0.675; OOB C-index, 0.679) (**Fig. 5e**). Integrating the two scores improved discrimination over either modality alone (OOB AUC, 0.770; OOB C-index, 0.739) (**Fig. 5e**; time-dependent AUC at 2, 3, and 5 years in **Supplementary Fig. S4b**). Consistent with these improvements, Kaplan–Meier plot showed that the combined score produced the clearest separation in recurrence-free survival compared with stratification based on either the H&E RRS or the tumor composite alone (**Fig. 5b, 5f,** and **Supplementary Fig. S4c**). Together, these results show that AI-defined histologic risk and spatially derived molecular signatures capture complementary dimensions of TNBC recurrence biology and that integrating them enables stronger, more temporally consistent prognostic stratification than either modality alone.

## Discussion

In this study, we developed an outcome-informed, AI-guided deep visual proteomics framework that uses H&E-derived recurrence-risk heatmaps to guide physical tissue sampling and MS-based proteomic profiling in TNBC. The central advance is the use of an outcome-guided model to determine where molecular investigation should take place, rather than treating its spatial predictions solely as an interpretive output. Within the same tumor compartment, high- and low-risk patches coexisted and showed distinguishable nuclear and architectural features. Exploratory profiling of 46 regions from two selected recurrent cases linked these contrasting image-derived risk states to mitotic programs in high-risk regions and immune-associated programs in low-risk regions. Together, these observations reveal recurrence-associated morpho-molecular signatures, connecting outcome-associated image signals with local histomorphology and directly measured molecular states. Concordant regional protein differences were used to construct a 13-protein composite for patient-level evaluation. Integrating this composite with the H&E-derived RRS yielded higher prognostic discrimination than the RRS alone in the expanded cohort, supporting a route from outcome-informed regional sampling to molecular discovery and patient-level evaluation of complementary prognostic information.

As the foundation of this framework, the patch-level model showed that aggregating the distribution of localized high-risk patches captured recurrence-associated signals more effectively than whole-slide averaging predictions within our cohort (**Fig. 1d–1e**). This aligns with the biological premise that recurrence may be driven by localized TME niches rather than by uniform properties of the whole tumor mass[31,32]. This selective emphasis on high-scoring regions broadly parallels the region-focused nature of histopathologic assessment while providing a computational approach to preserving their recurrence-associated score distribution. Furthermore, the patch-level design produced a spatially resolved risk map that could be directly leveraged for downstream molecular interrogation, linking patient-level prognostic modeling to region-level experimental sampling. Importantly, the principal value of the patch-level RRS in this work lies less in its standalone predictive accuracy than in the spatially resolved risk heatmap it generates, which provided the foundation for the core biological and spatial proteomic analyses central to this study.

At the molecular level, TNBC heterogeneity encompasses diverse tumor cell states, tissue compositions, and microenvironmental contexts, among which tumor-intrinsic proliferative, stromal or mesenchymal, and immune programs have emerged as recurrent organizing axes. Burstein et al.[33] found that the basal-like immune-suppressed subtype, characterized by proliferative activity and reduced immune signaling, exhibited the poorest disease-free and disease-specific survival, whereas the basal-like immune-activated subtype showed the most favorable outcomes. Lehmann et al.[34] subsequently confirmed these distinctions at the multi-omics level, demonstrating differential enrichment of cell-cycle, epithelial-mesenchymal transition, integrin, and immune and antigen presentation programs across the four TNBC subtypes. Collectively, previous studies have positioned immune-suppressed proliferative states toward the unfavorable extreme and immune-activated states toward the favorable extreme, whereas the prognostic significance of stromal, mesenchymal, and growth factor programs has been inconsistently explored[1,2,35,36]. Consistent with this framework, our bulk proteomic analysis identified a cohort-specific risk gradient, with cell cycle and genome maintenance programs enriched in the high-risk group; fibroblast proliferation, FGFR signaling, oxidative-stress responses, and angiogenic programs in the mid-risk group; and lymphocyte activation and antigen presentation programs in the low-risk group (**Fig. 2b–2e**). Together, these findings indicate that the AI-derived risk stratification reflects biologically distinct molecular states at the patient level, providing a basis for examining the local histomorphological manifestations of these states.At the histomorphological level, patch-level analysis showed that RRS was not simply a surrogate for tissue compartment identity. This was most evident within the tumor compartment, where patches spanned both extremes of the risk-score distribution and exhibited a non-unimodal pattern (**Fig. 3d–3g**). The contrasting nuclear and architectural features observed across the risk spectrum suggest that the model captured localized differences in tumor differentiation within the same histological compartment. These features broadly resemble nuclear pleomorphism and tubule formation, which are established components of conventional histological grading[37]. A recent computational pathology study linked quantitative nuclear morphometry in breast cancer to genomic instability, including aneuploidy, homologous-recombination deficiency, and whole-genome doubling. Although the specific nuclear features differed from those assessed here, this finding supports the broader possibility that the localized nuclear and architectural patterns captured by our model reflect biologically meaningful tumor states[38]. We also observed qualitative differences within the other tissue compartments. High-risk TAS patches tended to show denser and more sclerotic stroma, whereas low-risk immune patches showed a more aggregated spatial distribution of immune cells (**Supplementary Fig. 2f–2g**). Because TAS patches were distributed across the risk spectrum and immune patches were rare, we did not examine these non-tumor patterns in depth. Thus, they should be regarded as exploratory qualitative observations. Nevertheless, they are broadly consistent with previous studies showing that fibrotic stromal states and the spatial localization of immune cells can carry prognostic information in TNBC[1,2,39,40]. Together, these findings indicate that the AI-derived patch-level RRS captured localized variation in tumor cell morphology and microenvironmental organization beyond broad tissue compartment identity. The clear coexistence of contrasting high- and low-risk regions within the tumor compartment motivated subsequent spatial proteomic profiling of AI-defined high- and low-risk tumor regions.

Building on these histomorphological observations, we then used spatial omics to investigate the molecular states underlying the AI-defined high- and low-risk tumor regions. Among spatial omics technologies, spatial transcriptomics provides broad maps of gene expression programs and cellular niches, whereas multiplexed protein imaging platforms such as CODEX or PhenoCycler enable single-cell characterization of predefined proteins and spatial neighborhoods[41]. However, local transcript abundance does not necessarily correspond to protein abundance because of post-transcriptional regulation, protein turnover, and intracellular trafficking. Furthermore, multiplexed imaging remains constrained by the antibody panel composition[10,12]. MS-based proteomics provides a complementary strategy for broad, antibody-independent measurement of proteins from spatially selected tissue regions. For example, Mund et al.[17] combined AI-based image segmentation and phenotypic classification with laser microdissection and ultrasensitive MS-based proteomics to link spatially defined cellular phenotypes and disease states to their proteomic profiles. Similarly, our AI-guided deep visual proteomic framework translated image-derived information into physical tissue coordinates for molecular analysis However, the key distinction lay in the basis for defining the ROIs. Whereas Mund et al. selected cells or regions based on observable cell identity, morphology, marker expression, or disease stage, we defined our regions using RRSs derived directly from patient outcomes and localized within the same tumor compartment. The resulting risk heatmaps therefore functioned as outcome-informed sampling maps for isolating and molecularly interrogating AI-defined high- and low-risk tumor regions.

Using this spatial sampling strategy, spatial proteomics identified a directionally concordant molecular contrast across the profiled recurrence cases. Nineteen overlapping proteins showed concordant regulation, with high-risk tumor regions enriched in mitotic programs and low-risk regions enriched in immune- and antigen presentation–associated programs (**Fig. 4d–4e**). Because both high- and low-risk regions were sampled from tumor compartments, this contrast does not simply reflect differences between tumor and immune tissue compartments, but instead suggests that low-risk tumor regions were associated with a more immune-engaged local molecular state. At a different spatial scale, this regional contrast paralleled the proliferative-to-immune gradient observed in bulk proteomic analysis, providing cross-scale molecular support for the biological relevance of the AI-derived risk states. A collagen and stromal remodeling program was further observed in the high-risk regions of P1, suggesting that the shared proliferative state may be accompanied by patient-specific microenvironmental features[31]. Among the concordant proteins, the high-risk–associated P4HA1 has been linked to collagen remodeling, HIF-1α stabilization, treatment resistance, and unfavorable outcome in TNBC[42], whereas low-risk-associated proteins, including LCP1, PSME2, SERPINB9, and WAS, were consistent with leukocyte-associated and antigen-processing activities within tumor regions[1,2,43,44]. These findings indicate that AI-defined morphological risk corresponds to spatially localized proteomic states and support outcome-guided tissue sampling as a strategy for investigating the molecular basis of prognostic image signals.

To examine whether the spatially discovered tumor program could be translated to cohort-scale evaluation, we reformulated the concordant regional markers as a targeted bulk-tissue composite. Although bulk measurement necessarily loses the spatial organization of the original signal, the preservation of the spatially assigned risk orientation at the patient level, together with the concordant prognostic direction of the corresponding transcript-based composite in the METABRIC cohort[20,21], suggests that the regional high- versus low-risk contrast reflects a recurrent molecular axis rather than an idiosyncratic feature of the sampled coordinates. This spatial-to-bulk translation is conceptually consistent with previous breast cancer studies showing that locally organized tumor and microenvironmental states can provide recoverable tissue-level signatures in larger cohorts[45,46]. Image and molecular readouts showed differing trends in time-dependent discrimination (Supplementary Fig. S4b), broadly consistent with previous reports that the prognostic relevance of breast cancer features can vary over time[45,47-49]. The stronger discrimination of the H&E score at earlier evaluated time points can be considered in the context of the characteristic early recurrence pattern in TNBC[50] and previous reports of time-varying prognostic associations of proliferative activity and histological grade[47-49]. In contrast, discrimination estimates for the protein composite increased at later evaluated time points within the observed follow-up window. Related studies have identified molecular signatures associated with later recurrence, including epithelial–mesenchymal and stromal programs across breast cancer subtypes[45] and gene-expression assays in hormone receptor–positive disease[48,49]. These observations are compatible with the possibility of time-dependent complementarity: the H&E score may emphasize proliferative and architectural features relevant to nearer-term recurrence, whereas the spatially derived composite may capture additional molecular variation not fully represented by baseline morphology.

Several limitations should be considered in the context of the study design. First, deep MS-based proteomic profiling of small FFPE regions remains technically demanding with relatively low throughput. Therefore, spatial discovery was limited to two selected recurrence cases and was intended primarily to demonstrate the link between AI-defined risk regions and interpretable molecular states, rather than to estimate the population-level prevalence of these states. Nevertheless, multiple high- and low-risk regions were profiled within each case, and concordant signals were subsequently evaluated in larger protein- and transcript-level cohorts. Future workflows using larger contiguous regions defined directly by the risk heatmaps may improve material yield and facilitate more systematic patient-level and multi-region profiling. Spatial analysis was initially focused on the tumor compartment because it contained sufficiently represented high- and low-risk regions for within-compartment comparison. However, the prognostic relevance of stromal and immune organization, along with the risk-associated distributions observed in our TAS and immune patches, supports extending heatmap-guided spatial profiling to these latter two compartments. Second, cohort-scale evaluation relied on bulk FFPE proteomics, which cannot preserve the original spatial arrangement or fully resolve cellular sources. However, it provided a deliberate and scalable means of testing whether localized molecular programs leave a reproducible patient-level footprint. Third, the modest number of recurrence events limits the precision of the performance estimates, particularly at individual follow-up time points. The observed temporal patterns remain exploratory and do not establish distinct biological mechanisms of early and later recurrence. Independent TNBC cohorts with sufficient recurrence events and follow-up are needed to assess the reproducibility of these patterns and whether the incremental prognostic value of the composite varies with the prediction horizon. Finally, the identified regions and proteomic programs should be interpreted as recurrence-associated rather than causal. Higher-resolution spatial and functional studies will be needed to establish their cellular origins and mechanistic roles.

Despite these limitations, this study presents an outcome-informed, AI-guided deep visual proteomics framework that links H&E-based risk mapping to multiscale proteomic analysis. More fundamentally, our findings suggest that outcome-informed image models can resolve intratumoral mosaics of morphologically distinguishable tissue states and convert these states into spatially explicit units for molecular investigation. In this context, the recurrence-risk heatmap is no longer merely a predictive output but becomes an experimental layer that identifies sites of concentrated outcome-associated morphology and directs targeted tissue sampling. Coupling these image-defined units with spatial profiling provides a means to test whether prognostic morphology corresponds to reproducible tumor and microenvironmental programs, preserving intratumoral heterogeneity otherwise obscured by bulk measurements. With further methodological refinement, this strategy could enable region-aware, patient-specific biomarker systems that integrate information from spatially defined tumor, stromal, and immune states to inform recurrence-risk assessment, thereby advancing computational pathology beyond mere predictive utility toward a spatially and biologically grounded understanding of TNBC recurrence.

## Methods

### Study population and specimens

This retrospective study included 156 patients with surgically resected TNBC at Kangbuk Samsung Hospital between 2004 and 2024. Matched FFPE tissue specimens, H&E-stained WSIs, clinicopathological data, and recurrence follow-up data were collected. The primary cohort was divided into a development cohort (n = 107) and an test cohort (n = 49). We collected and included 47 additional patients of TNBCfor cohort-level evaluation of the spatial proteomics-derived signature, yielding an expanded cohort of 96 patients. The study was approved by the Institutional Review Board of Kangbuk Samsung Hospital (IRB No. 2024-08-043-001). Written informed consent was obtained from all participants.

### Model development and evaluation

Foreground tissue regions were identified on each WSI using a fixed-threshold segmentation of the saturation channel with morphological artifact removal and partitioned into non-overlapping 512 × 512-pixel patches at 40× magnification, which corresponds to approximately 150 × 150-μm patches. A tissue-compartment classifier initialized from a breast cancer–distilled ConvNeXt-Base model[51,52,53] was fine-tuned using the TIGER dataset[26] to classify patches as tumor, tumor-associated stroma, immune, necrosis, or rest. Patches classified as rest were excluded from recurrence-model development.

We trained patch-level recurrence classifier in the development cohort (n = 107) using weak supervision, where each retained patch inherited its patient's recurrence status. We fine-tuned the ConvNeXt-Base model using binary cross-entropy with label smoothing and balanced mini-batch sampling. The trained classifier generated an patch-level RRS between 0 and 1 for each patch. The patch-level RRSs were mapped to their original WSI coordinates to construct spatial recurrence-risk heatmaps.

To compare patient-level RRS strategies, we evaluated various combinations of patch selection (all patches or the highest-scoring patches) and score aggregation strategies (mean or histogram-based aggregation). The selected configuration represented the 10 highest-scoring patches using a 20-bin RRS histogram. We then applied Lasso regression to derive the patient-level RRS. Detailed procedures for WSI preprocessing, patch filtering, model training, optimization, and integration of other layers are provided in the **Supplementary Methods**.

**Proteomics analysis**

Bulk FFPE tissue samples and SLACS-isolated spatial regions were collected into Adaptive Focused Acoustics (AFA)-TUBE TPX Strips (Covaris) and processed using an sodium dodecyl sulfate (SDS)-assisted AFA workflow. Samples were lysed in 5% SDS and 50 mM triethylammonium bicarbonate (pH 8.5), subjected to AFA, heated at 90 °C for 1 h to reverse formaldehyde-induced cross-links, and subjected to a second AFA cycle. This procedure enabled protein extraction without organic-solvent deparaffinization. Clarified lysates were processed on an AssayMAP Bravo platform (Agilent Technologies), including reduction with dithiothreitol, alkylation with chloroacetamide, and digestion with RapiZyme (Waters). Peptides were separated using an Evosep Eno LC system and analyzed by data-independent acquisition (DIA)–MS on an Orbitrap Astral mass spectrometer (Thermo Fisher Scientific). Raw data were processed using DIA-NN (version 2.3.2) in library-free mode against the UniProt human reference proteome. Precursor and protein false discovery rates were controlled at ≤1%, and quantification was performed using retention-time-dependent cross-run normalization in QuantUMS in high-precision mode. Detailed methods on sample processing, acquisition, quantification, and data analysis are provided in the **Supplementary Methods**.

**AI-guided spatial tissue isolation via SLACS**

Spatial proteomic analysis was performed using adjacent FFPE sections from two patients with recurrence in the test cohort. Using image-derived criteria, we selected cases that represent the greatest within-slide variance in patch-level RRS and the highest patient-levelRRS. Serial 10-μm sections were cut from the corresponding FFPE blocks, mounted on CosmoSlides (Meteor Biotech), deparaffinized, post-fixed, stained with H&E, and digitally scanned. The resulting WSIs were co-registered with the corresponding AI-analyzed H&E images, patch-level RRS heatmaps, and TIGER-derived tissue compartment maps using AstroMapper software (Meteor Biotech).

Following image registration, candidate patches were manually annotated through joint review of the aligned H&E images, RRS heatmaps, and tissue compartment maps. Tumor patches were classified as tumor high-risk (T_high) or tumor low-risk (T_low) according to their patch-level RRSs. The coordinates of the selected 150-μm patches were imported into the CosmoSort spatial cell sorter (Meteor Biotech), and the corresponding tissue patches were isolated by SLACS (spatially resolved laser-activated cell sorting; Meteor Biotech)[30] into individually labeled collection tubes. To obtain sufficient material for proteomic analysis, two to six neighboring patches assigned to the same tissue compartment and risk category were pooled as a single spatial region. The collected regions were subsequently processed for LC–MS proteomic analysis. Details on tissue preparation, image registration, manual region selection, and SLACS-based[30] spatial isolation are provided in the **Supplementary Methods**.

**Statistical analysis**

Clinicopathological variables were compared using parametric or non-parametric tests for continuous variables and the $\chi^2$ test or Fisher's exact test for categorical variables, as appropriate. Departure from unimodality in patch-level RRSdistributions was assessed using Hartigan's dip test[29].

Recurrence-free survival was defined as the interval from surgery to recurrence or the last clinical follow-up. Survival associations were evaluated using Kaplan–Meier analysis, the log-rank test, and Cox proportional hazards regression. Hazard ratios were reported per 1 standard-deviation increase in continuous variables. Predictive performance was assessed using the AUC, C-index, and time-dependent AUCs at 2, 3, and 5 years. The H&E RRS, tumor composite, and integrated score were evaluated using 200 out-of-bag bootstrap iterations. All tests were two-sided, and $P < 0.05$ was considered statistically significant.

## Code availability

Code used for whole-slide image preprocessing, the tissue compartment and patch-level RRS generation, patient-level risk-score aggregation, and the targeted proteomic panel analysis is available at https://github.com/Radisen-Panthera/TNBC-Recurrence-Prediction-Official.

## Data availability

The clinical, pathological, and outcome data supporting the findings of this study, together with the corresponding H&E whole-slide images and proteomic data, are not publicly available because of patient privacy protections and the terms of the Institutional Review Board approval (Kangbuk Samsung Hospital, IRB No. 2024-08-043-001) and participant informed consent, which do not permit open deposition of patient-level data. These data are available from the corresponding author upon reasonable request, subject to institutional approval and execution of a data use/transfer agreement.

The TIGER challenge dataset used to train the tissue compartment classifier is publicly available at https://tiger.grand-challenge.org/ under a CC BY-NC 4.0 license. The METABRIC cohort data used to validate the prognostic direction of the transcript-level analog of the tumor composite score in an independent cohort are publicly available via cBioPortal (https://www.cbioportal.org).

There are no conflicts of interest to declare

## Acknowledgements

None of interest

## Funding

None of interest.

## Author contributions

Y.S.C let the study, developed the AI model and computational pipeline, and performed formal analysis of the imaging data. J.H.P. led the study and performed proteomic data analysis and interpretation. C.I.K. co-led the study and was responsible for proteomic data generation. H.W.K., Y.Y.L., H.S.R., S.M.L., and A.C.W.L. performed the SLCAS-based spatial tissue isolation, H.L.L., Y.M.L., and S.M.P. performed proteomic data analysis, G.G.L., S.J.H., C.H.L., S.H.K., and H.J.P., contributed to result interpretation. H.Y.S., H.W.K., H.J.S., S.H.P., J.H.Y., S.M.P., S.W.K., and Y.J.K. performed proteomic data generation. S.I.D provided pathological interpretation and advice. E.Y.K. provided clinical advice. D.M.S. and J.B.P. supervised this study. I.G.D. supervised the study and provided pathological interpretation and advice. All authors contributted to and approved the final manuscript.

## Competing interests

H.K., Y.Y.L., H.S.R., S.M.L. and A.C.W.L. are employees of Meteor Biotech Co. Ltd., which developed the spatial tissue isolation platform used in this study. The remaining authors declare no competing interests.

# Figure Legends

**Fig. 1. Triple-negative breast cancer (TNBC) recurrence prediction framework and performance of the hematoxylin and eosin (H&E)–based AI model. (a)** Overview of the patch-level recurrence prediction pipeline. H&E whole-slide images (WSIs) from the TNBC cohort were partitioned into 150-μm patches and subjected to in-house patch-quality filtering. Each retained patch was assigned a patch-level recurrence-associated risk score (RRS; 0–1), which was remapped to its original WSI coordinates to generate a spatial recurrence-risk heatmap. For patient-level prediction, patch-level RRSs were aggregated using a histogram representation of the top-scoring patches, followed by Lasso regression (Hist. + Top-k). **(b)** Patch-level distribution of RRSs according to patient recurrence

status in the test cohort. A Sankey plot shows the allocation of patches from patients with and without recurrence across five RRS bins. Pie charts indicate the proportions of recurrence and non-recurrence patches within each bin. **(c)** Patient-level distribution of patch RRSs. For each patient, the fraction of patches assigned to each RRS bin was calculated and averaged within the recurrent and non-recurrent groups. Bars indicate the mean within-patient patch fraction, and error bars indicate the standard error of the mean. **(d)** Receiver operating characteristic curves comparing four patient-level aggregation strategies based on histogram versus mean-pool aggregation and top-k patch selection versus the use of all retained patches in the test cohort. **(e)** Comparison of concordance indices between the Hist. + Top-k model and representative slide-level weakly supervised multiple-instance learning models, including CLAM_mb, CLAM_sb, ABMIL, and TransMIL, in the test cohort.

**Fig. 2. Bulk proteomic characterization of AI-derived risk groups in triple-negative breast cancer (TNBC). (a)** Schematic overview of the bulk proteomic workflow. Formalin-fixed paraffin-embedded (FFPE) TNBC tissue specimens were subjected to liquid chromatography–tandem mass spectrometry (LC-MS/MS)-based data-independent acquisition (DIA) proteomic analysis. **(b)** Heatmaps showing the abundance patterns of differentially expressed proteins (DEPs) identified in the high-risk (n = 8), mid-risk (n = 9), and low-risk (n = 24) groups. DEPs were classified according to preferential upregulation in the high-risk (high-up; n = 118), mid-risk (mid-up; n = 44), or low-risk (low-up; n = 77) group. Protein abundance is expressed as row-wise z-scores. **(c)** Functional enrichment analysis of DEPs preferentially upregulated in each risk group. Selected enriched biological processes are shown, with color intensity representing −log10(P). **(d)** Comparison of ESTIMATE-derived immune scores between the low-risk group, designated as the target group, and the remaining patients. P values are indicated above the comparisons. **(e)** Comparison of ESTIMATE-derived tumor purity scores between the high-risk group, designated as the target group, and the remaining patients. P values are indicated above the corresponding comparisons.

**Fig. 3. Tissue-compartment distribution and morphological heterogeneity of AI-derived patch-level risk scores in triple-negative breast cancer (TNBC). (a)** UMAP projection of DINO AI-derived patch features colored by recurrence-associated risk score (RRS). **(b)** UMAP projection of the same patches colored by tissue compartment, including tumor, tumor-associated stroma (TAS), necrosis, and immune compartments. **(c)** Sankey plot showing patch allocation from each tissue compartment across five RRS bins. Pie charts indicate the relative tissue-compartment composition of the patches assigned to each RRS bin. **(d)** Violin plots showing the distribution of patch-level RRS values across the four tissue compartments. **(e)** Distribution of RRS values among tumor patches. Gray lines indicate the observed score distribution, and the black curve represents the kernel density estimate.

Departure from unimodality was assessed using Hartigan's dip test. **(f)** Violin plots showing the upper and **(g)** lower quartiles of RRS values within each tissue compartment. **(h)** UMAP projection of tumor patches colored by patch-level RRS, with the locations of patches selected for histopathologic review indicated (left). Among the 10 highest- and 10 lowest-scoring tumor patches examined, a representative subset of five patches from each risk group is shown in the main figure (right).

**Fig. 4. Spatial proteomic profiling of AI-defined high- and low-risk tumor regions in triple-negative breast cancer (TNBC). (a)** Schematic overview of the spatial proteomics workflow. Regions of interest (ROIs) were selected using AI-derived recurrence-risk heatmaps, isolated by infrared pulse laser–based cutting, collected, and subjected to LC–MS/MS-based data-independent acquisition (DIA) proteomic analysis. **(b)** Distribution of tumor and immune patch-level recurrence-associated risk scores (RRSs) across recurrent cases with available adjacent formalin-fixed paraffin-embedded sections. Box plots show the patch-level score distributions, and the black points connected by a dashed line indicate the corresponding patient-level RRSs. The two patients (P1 and P2) selected for spatial proteomic profiling are highlighted in orange. **(c)** Representative tumor patch RRS heatmap from P1, showing the locations of high- and low-risk tumor regions selected for spatial proteomic profiling. Enlarged images showing representative groups of neighboring patches collected from the high- and low-risk regions. Scale bar, 270 μm. **(d)** Functional enrichment analysis of differentially expressed proteins (DEPs) upregulated in high- or low-risk tumor regions from P1 and P2. Selected enriched biological processes are displayed (color intensity representing −log10[P]). **(e)** Box plots showing the standardized abundance of 19 DEPs in both patients with concordant directions of regulation. For each protein, abundance values were standardized separately within each patient using the corresponding low-risk regions as the reference, such that the low-risk group was centered at zero. Proteins preferentially expressed in high- and low-risk tumor regions are indicated by magenta and green labels, respectively. Box fill colors indicate patch risk group; box outlines indicate the patient.

**Fig. 5. Cohort-level evaluation of the spatial proteomics-derived tumor composite and hematoxylin and eosin (H&E)–based AI risk score in triple-negative breast cancer (TNBC). (a)** Schematic overview of the expanded-cohort evaluation. The original test cohort (n = 49) was expanded to include 47 additional patients, totaling 96 patients with matched formalin-fixed paraffin-embedded tissue specimens and H&E whole-slide images. DIA-MS data were analyzed by target post-acquisition analysis to quantify 33 precursors representing 13 spatial proteomics-derived candidate proteins. One representative precursor per protein was selected to construct the 13-

protein tumor composite. In parallel, patient-level H&E RRSs were generated using the patch-level recurrence-risk model. The tumor composite was subsequently added to the H&E RRSs to evaluate the integrative model. **(b)** Kaplan–Meier curves for recurrence-free survival (RFS) in the expanded cohort stratified by tumor composite (n = 96; 12 recurrence events). **(c)** Kaplan–Meier curves for RFS stratified by corresponding transcript-based composite in the METABRIC TNBC cohort (n = 261; 103 recurrence events). **(d)** Forest plots of the univariate Cox proportional hazards analyses for the composite and its individual markers in the expanded and METABRIC cohorts. Hazard ratios are expressed as per 1-standard-deviation increase after aligning marker directions according to their spatially defined risk orientation. Upward and downward arrows indicate proteins preferentially expressed in high-risk and low-risk tumor regions, respectively. Circles with solid confidence intervals represent the expanded cohort; diamonds with dotted confidence intervals represent the METABRIC cohort. Blue symbols indicate hazard ratio directions consistent with the spatially defined risk orientation; gray symbols indicate reversed directions; red symbols indicate the composite. *$P < 0.05$. **(e)** 5-year ROC curves out-of-bag (OOB) bootstrap for the H&E AI-derived patient-level RRS, the tumor composite, and the combined RRS + composite model in the expanded cohort, with corresponding OOB AUC and C-index for each score. (f) Kaplan-Meier curves for RFS stratified by combined RRS + composite score in the expanded cohort (n = 96; 12 recurrence events).

# Table

Table 1. Clinicopathological characteristics of TNBC cohorts

| | Developing cohort (n = 107) | Test cohort (n = 49) | p-value |
|---|---|---|---|
| Recurrence, n (%) | 13 (12.1%) | 7 (14.3%) | 0.797 ‡ |
| Time-to-recurrence, months (mean ± SD) * | 12.4 ± 8.1 | 16.6 ± 8.6 | 0.275 † |
| Age, years (mean ± SD) | 56.5 ± 10.6 | 55.0 ± 13.3 | 0.435 † |
| **Histology type, n (%)** | | | 0.260 § |
| Invasive ductal carcinoma | 89 (83.2%) | 44 (89.8%) | |
| Invasive lobular carcinoma | 4 (3.7%) | 1 (2.0%) | |
| Metaplastic carcinoma | 7 (6.5%) | 1 (2.0%) | |
| Medullary carcinoma | 0 (0.0%) | 0 (0.0%) | |
| Mucinous | 0 (0.0%) | 1 (2.0%) | |
| Micropapillary | 1 (0.9%) | 0 (0.0%) | |
| Other/Special type | 6 (5.7%) | 2 (4.1%) | |
| **Histologic grade, n (%)** | | | 0.602 § |
| Grade 1 | 8 (7.5%) | 3 (6.1%) | |
| Grade 2 | 37 (34.6%) | 14 (28.6%) | |
| Grade 3 | 60 (56.1%) | 32 (65.3%) | |
| Not evaluable | 2 (1.9%) | 0 (0.0%) | |
| Tumor size of invasive tumor, mm (mean ± SD) | 23.0 ± 17.8 | 28.2 ± 14.0 | 0.002 † |
| **Lymphovascular invasion, n (%)** | | | 0.246 § |
| Yes | 24 (22.4%) | 16 (32.7%) | |
| No | 83 (77.6%) | 33 (67.3%) | |
| **T stage, n (%)** | | | 0.014 § |
| T1 | 64 (59.8%) | 17 (34.7%) | |
| T2 | 33 (30.8%) | 28 (57.1%) | |
| T3 | 9 (8.4%) | 3 (6.1%) | |
| T4 | 1 (0.9%) | 1 (2.0%) | |
| **N stage, n (%)** | | | 0.822 § |
| N0 | 72 (67.3%) | 30 (61.2%) | |
| N1 | 21 (19.6%) | 9 (18.4%) | |
| N2 | 8 (7.5%) | 6 (12.2%) | |
| N3 | 5 (4.7%) | 3 (6.1%) | |
| Unknown | 1 (0.9%) | 1 (2.0%) | |

Data are presented as mean ± standard deviation or n (%).

* Time-to-recurrence is calculated only for patients who experienced recurrence (Image Developing: n = 13; Test: n = 7).

† Mann-Whitney U test.

‡ Fisher's exact test.

§ Chi-square test.

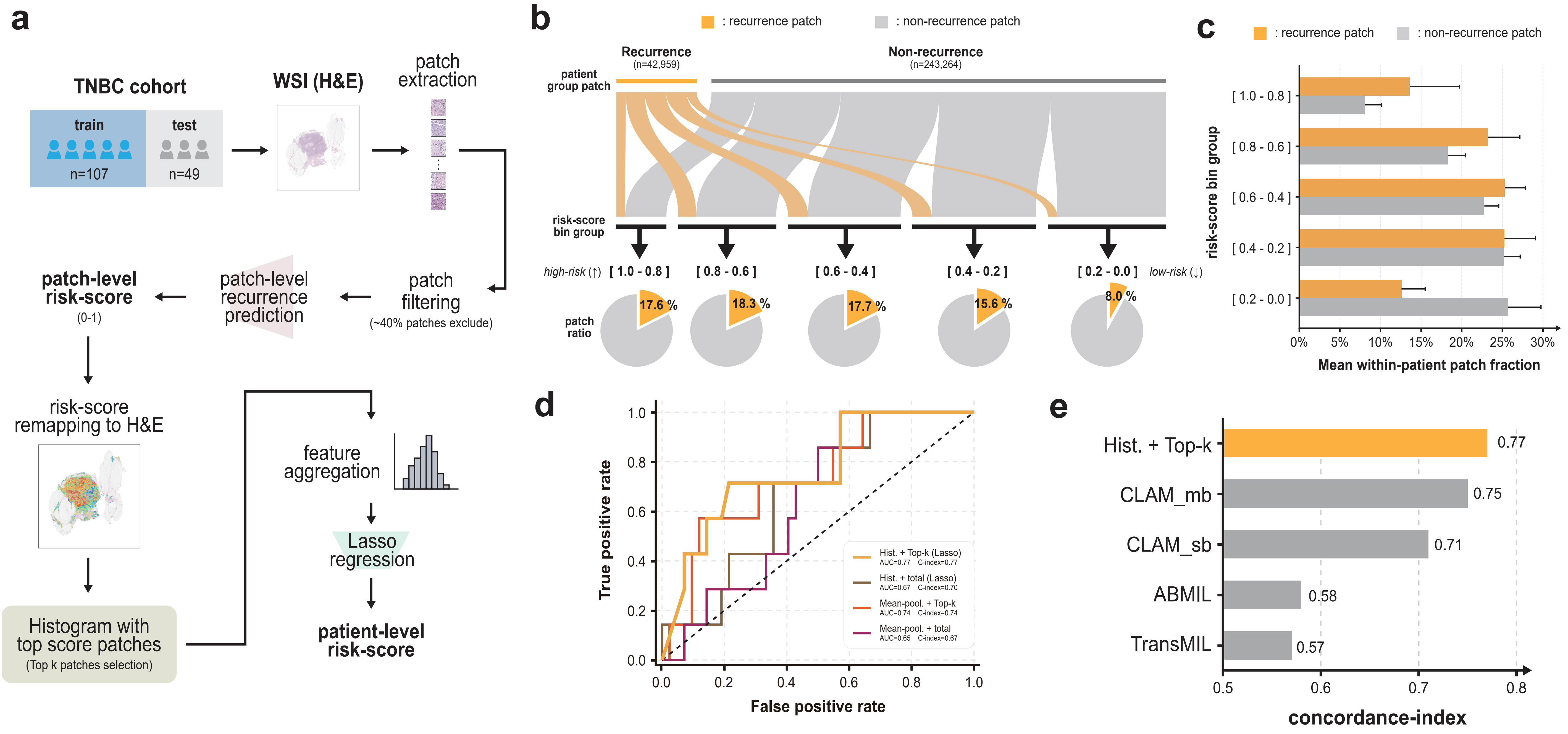

a
TNBC cohort
train
n=107
test
n=49
WSI (H&E)
patch extraction
patch filtering
(~40% patches exclude)
patch-level recurrence prediction
patch-level risk-score
(0-1)
risk-score remapping to H&E
Histogram with top score patches
(Top k patches selection)
feature aggregation
Lasso regression
patient-level risk-score
b
: recurrence patch
: non-recurrence patch
Recurrence
(n=42,959)
Non-recurrence
(n=243,264)
patient group patch
risk-score bin group
high-risk (↑)
[ 1.0 - 0.8 ]
[ 0.8 - 0.6 ]
[ 0.6 - 0.4 ]
[ 0.4 - 0.2 ]
[ 0.2 - 0.0 ]
low-risk (↓)
patch ratio
17.6 %
18.3 %
17.7 %
15.6 %
8.0 %
c
: recurrence patch
: non-recurrence patch
risk-score bin group
[ 1.0 - 0.8 ]
[ 0.8 - 0.6 ]
[ 0.6 - 0.4 ]
[ 0.4 - 0.2 ]
[ 0.2 - 0.0 ]
0%
5%
10%
15%
20%
25%
30%
Mean within-patient patch fraction
d
True positive rate
False positive rate
0.0
0.2
0.4
0.6
0.8
1.0
Hist. + Top-k (Lasso)
AUC=0.77 C-index=0.77
Hist. + total (Lasso)
AUC=0.67 C-index=0.70
Mean-pool. + Top-k
AUC=0.74 C-index=0.74
Mean-pool. + total
AUC=0.65 C-index=0.67
e
Hist. + Top-k
0.77
CLAM_mb
0.75
CLAM_sb
0.71
ABMIL
0.58
TransMIL
0.57
0.5
0.6
0.7
0.8
concordance-index

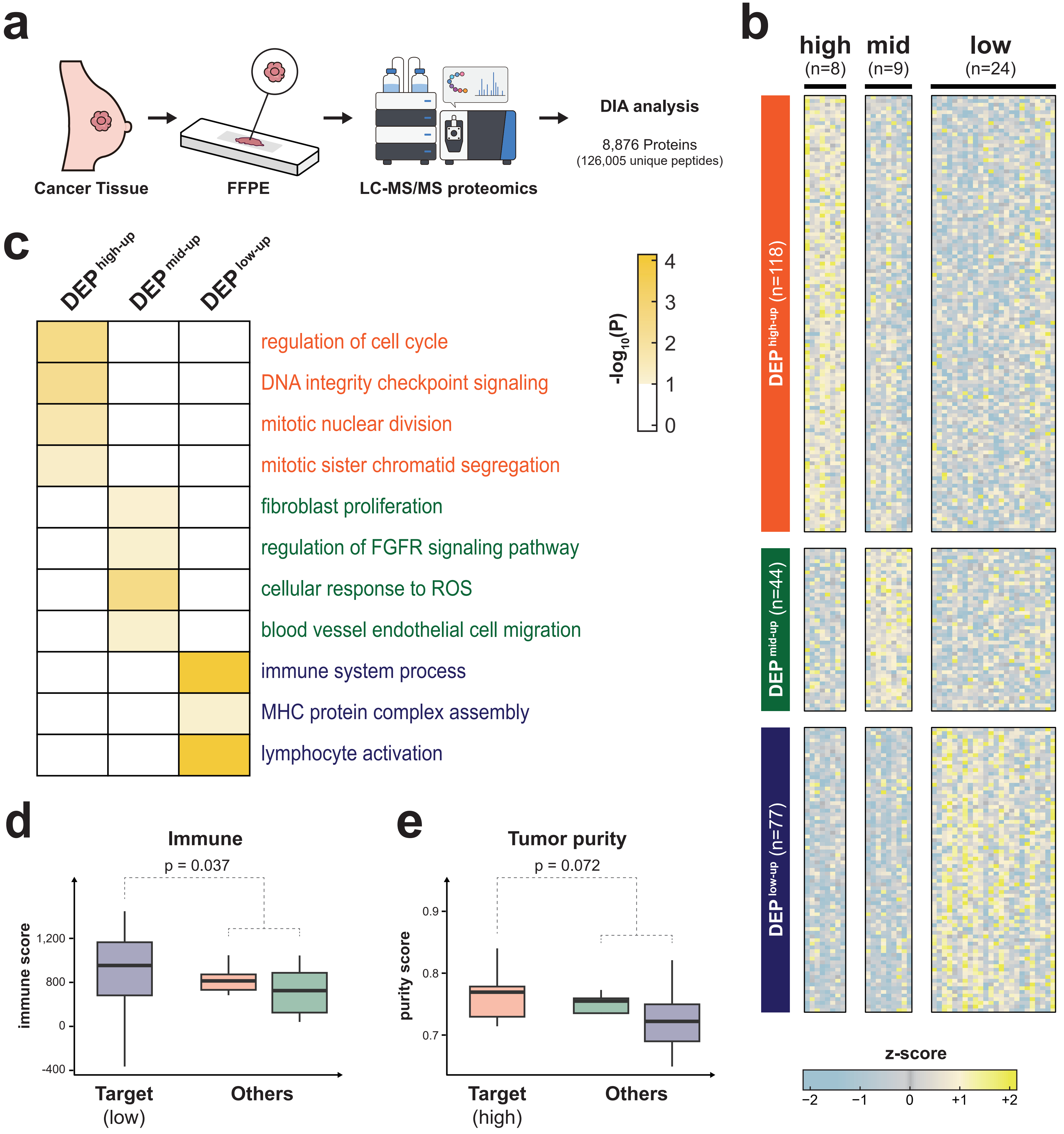
a
Cancer Tissue
FFPE
LC-MS/MS proteomics
DIA analysis
8,876 Proteins
(126,005 unique peptides)
b
high
(n=8)
mid
(n=9)
low
(n=24)
DEP high-up (n=118)
DEP mid-up (n=44)
DEP low-up (n=77)
z-score
−2
−1
0
+1
+2
c
DEP high-up
DEP mid-up
DEP low-up
regulation of cell cycle
DNA integrity checkpoint signaling
mitotic nuclear division
mitotic sister chromatid segregation
fibroblast proliferation
regulation of FGFR signaling pathway
cellular response to ROS
blood vessel endothelial cell migration
immune system process
MHC protein complex assembly
lymphocyte activation
-log10(P)
4
3
2
1
0
d
Immune
p = 0.037
immune score
1,200
800
0
-400
Target
(low)
Others
e
Tumor purity
p = 0.072
purity score
0.9
0.8
0.7
Target
(high)
Others

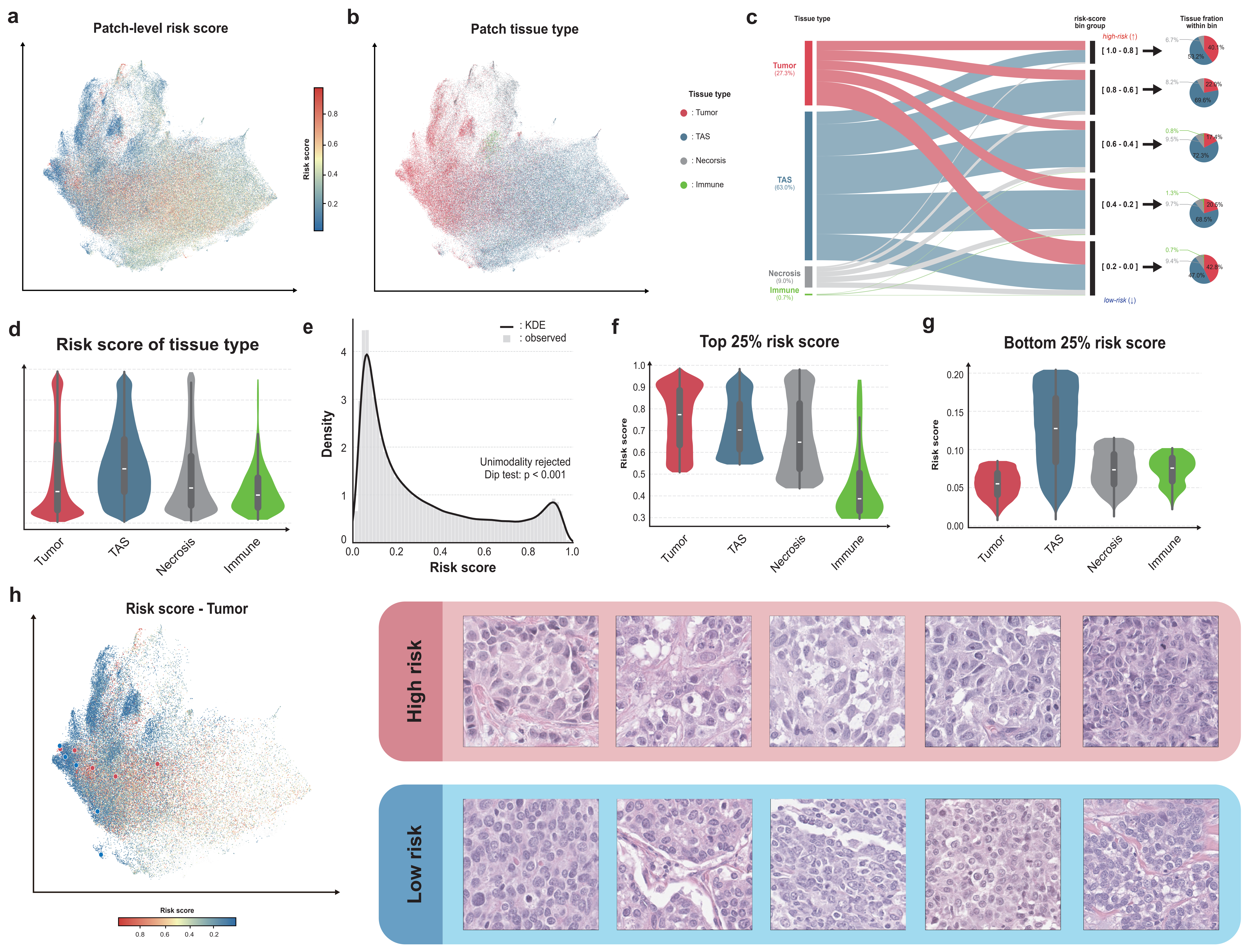

a
Patch-level risk score
Risk score
0.8
0.6
0.4
0.2
b
Patch tissue type
Tissue type
: Tumor
: TAS
: Necorsis
: Immune
c
Tissue type
risk-score bin group
Tissue fration within bin
Tumor (27.3%)
TAS (63.0%)
Necrosis (9.0%)
Immune (0.7%)
high-risk (↑)
[ 1.0 - 0.8 ]
[ 0.8 - 0.6 ]
[ 0.6 - 0.4 ]
[ 0.4 - 0.2 ]
[ 0.2 - 0.0 ]
low-risk (↓)
d
Risk score of tissue type
Tumor
TAS
Necrosis
Immune
e
: KDE
: observed
Density
Unimodality rejected
Dip test: p < 0.001
Risk score
f
Top 25% risk score
Risk score
g
Bottom 25% risk score
Risk score
h
Risk score - Tumor
High risk
Low risk
Risk score

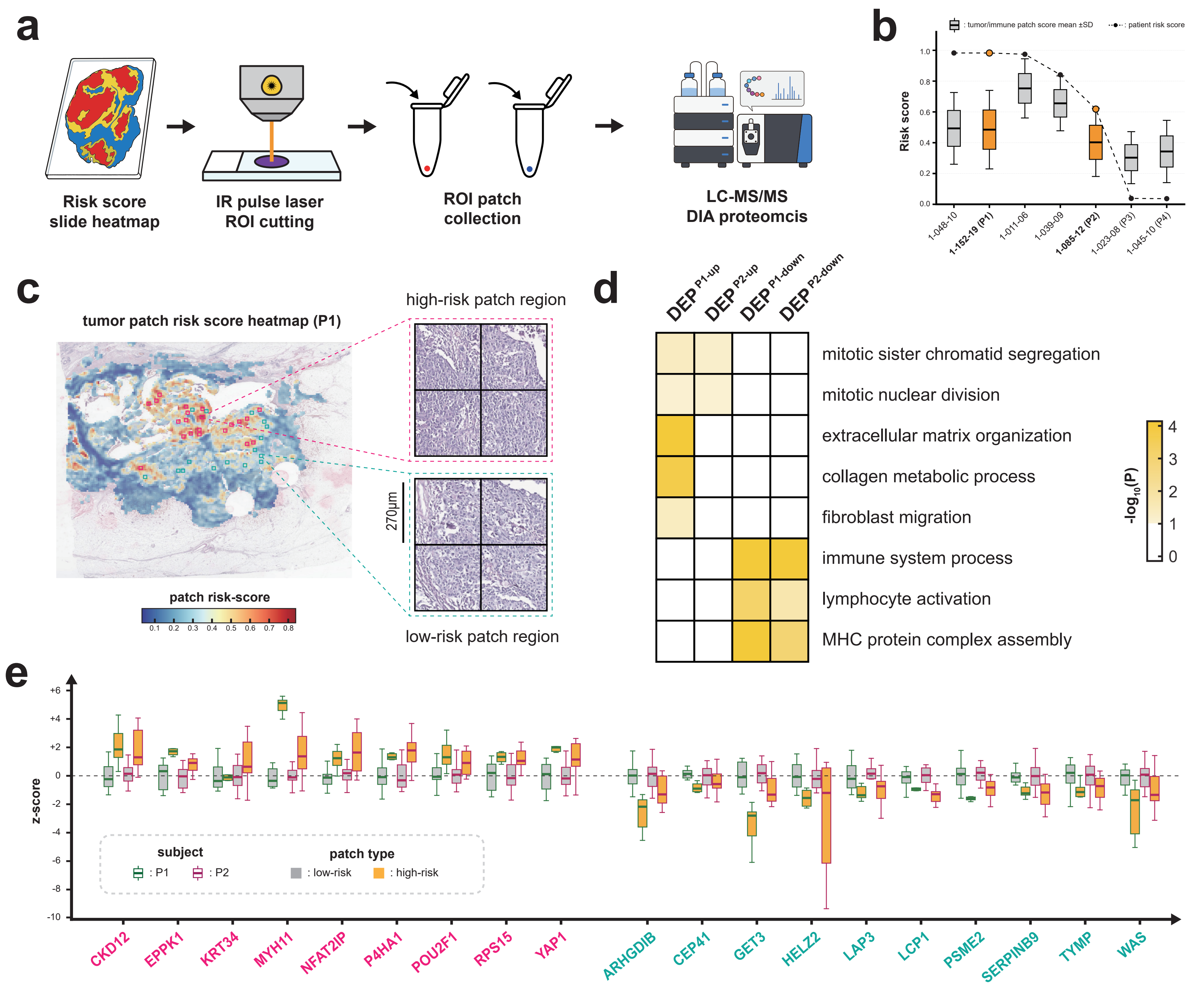

a
Risk score slide heatmap
IR pulse laser ROI cutting
ROI patch collection
LC-MS/MS DIA proteomcis
b
: tumor/immune patch score mean ±SD
: patient risk score
Risk score
1-048-10
1-152-19 (P1)
1-011-06
1-039-09
1-085-12 (P2)
1-023-08 (P3)
1-045-10 (P4)
c
tumor patch risk score heatmap (P1)
high-risk patch region
270μm
low-risk patch region
patch risk-score
d
DEP P1-up
DEP P2-up
DEP P1-down
DEP P2-down
mitotic sister chromatid segregation
mitotic nuclear division
extracellular matrix organization
collagen metabolic process
fibroblast migration
immune system process
lymphocyte activation
MHC protein complex assembly
-log10(P)
e
z-score
subject
: P1
: P2
patch type
: low-risk
: high-risk
CKD12
EPPK1
KRT34
MYH11
NFAT2IP
P4HA1
POU2F1
RPS15
YAP1
ARHGDIB
CEP41
GET3
HELZ2
LAP3
LCP1
PSME2
SERPINB9
TYMP
WAS

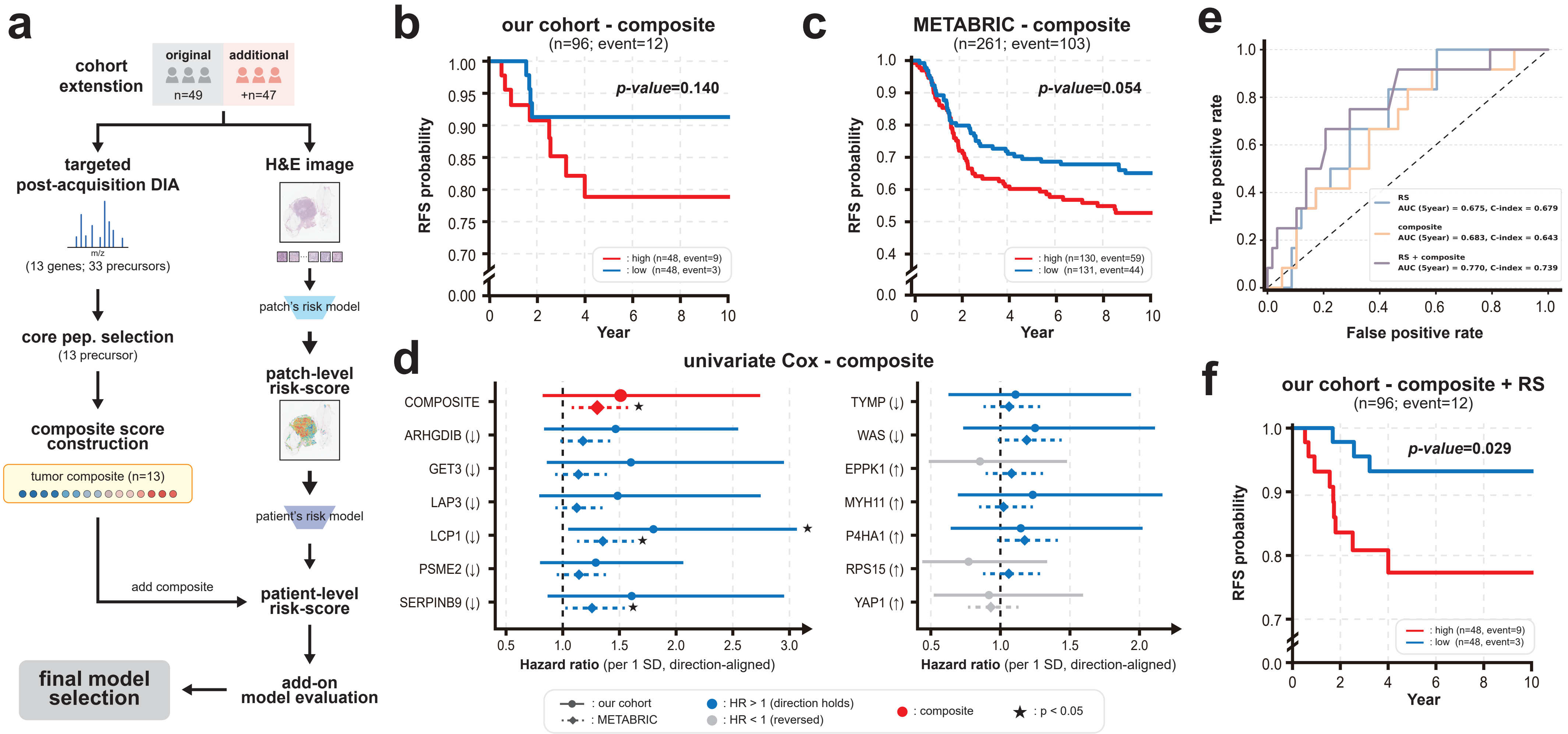
a
cohort extenstion
original n=49
additional +n=47
targeted post-acquisition DIA
m/z
(13 genes; 33 precursors)
core pep. selection
(13 precursor)
composite score construction
tumor composite (n=13)
add composite
H&E image
patch's risk model
patch-level risk-score
patient's risk model
patient-level risk-score
add-on model evaluation
final model selection
b
our cohort - composite
(n=96; event=12)
p-value=0.140
RFS probability
Year
: high (n=48, event=9)
: low (n=48, event=3)
c
METABRIC - composite
(n=261; event=103)
p-value=0.054
RFS probability
Year
: high (n=130, event=59)
: low (n=131, event=44)
e
True positive rate
False positive rate
RS
AUC (5year) = 0.675, C-index = 0.679
composite
AUC (5year) = 0.683, C-index = 0.643
RS + composite
AUC (5year) = 0.770, C-index = 0.739
d
univariate Cox - composite
COMPOSITE
ARHGDIB (↓)
GET3 (↓)
LAP3 (↓)
LCP1 (↓)
PSME2 (↓)
SERPINB9 (↓)
TYMP (↓)
WAS (↓)
EPPK1 (↑)
MYH11 (↑)
P4HA1 (↑)
RPS15 (↑)
YAP1 (↑)
Hazard ratio (per 1 SD, direction-aligned)
: our cohort
: METABRIC
: HR > 1 (direction holds)
: HR < 1 (reversed)
: composite
: p < 0.05
f
our cohort - composite + RS
(n=96; event=12)
p-value=0.029
RFS probability
Year
: high (n=48, event=9)
: low (n=48, event=3)